\documentclass[letterpaper, 10pt, conference]{ieeeconf}      

\IEEEoverridecommandlockouts                              

\usepackage[utf8]{inputenc}
\usepackage[english]{babel}

\usepackage{amsmath}
\usepackage{multicol}
\usepackage{mathtools}
\usepackage[table]{xcolor}
\usepackage[verbose]{wrapfig}
\usepackage{url}
\usepackage{svg}
\usepackage{caption}
\usepackage{threeparttable}
\usepackage{graphicx}
\usepackage{array}
\usepackage{hhline}
\usepackage{breqn}
\newcolumntype{P}[1]{>{\centering\arraybackslash}p{#1}}
 
\title{\LARGE \bf
Revisiting Random Forests in a Comparative Evaluation of Graph Convolutional Neural Network Variants for Traffic Prediction*
}

\author{Ta Jiun Ting$^{1}$, Xiaocan Li$^{1}$, Scott Sanner$^{1}$, and Baher Abdulhai$^{2}$
\thanks{*The authors would like to thank Huawei Canada Research Centre for financial \& technical support.}
\thanks{$^{1}$Ta Jiun Ting, Xiaocan Li, and Scott Sanner are with the Department of Mechanical \& Industrial Engineering, University of Toronto, 5 King’s College Road, Toronto, Ontario, M5S 3G8, Canada
        {\tt\small tajiun.ting@mail.utoronto.ca, hsiaotsan.li@mail.utoronto.ca, ssanner@mie.utoronto.ca}}%
\thanks{$^{2}$Baher Abdulhai is with the Department of Civil \& Mineral Engineering, University of Toronto, 35 St. George Street, Toronto, Ontario, M5S 1A4, Canada
        {\tt\small baher.abdulhai@utoronto.ca}}%
}

\begin{document}

\maketitle
\begin{abstract}
Traffic prediction is a spatiotemporal predictive task that plays an essential role in intelligent transportation systems. Today, graph convolutional neural networks (GCNNs) have become the prevailing models in the traffic prediction literature since they excel at extracting spatial correlations. In this work, we classify the components of successful GCNN prediction models and analyze the effects of matrix factorization, attention mechanism, and weight sharing on their performance. Furthermore, we compare these variations against random forests, a traditional regression method that predates GCNNs by over 15 years. We evaluated these methods using simulated data of two regions in Toronto as well as real-world sensor data from selected California highways. We found that incorporating matrix factorization, attention, and location-specific model weights either individually or collectively into GCNNs can result in a better overall performance. Moreover, although random forest regression is a less compact model, it matches or exceeds the performance of all variations of GCNNs in our experiments. This suggests that the current graph convolutional methods may not be the best approach to traffic prediction and there is still room for improvement. Finally, our findings also suggest that for future research on GCNN for traffic prediction to be credible, researchers must include performance comparison to random forests.
\end{abstract}

\section{Introduction}
Accurate traffic prediction is an integral component of intelligent transportation systems (ITS) as it is critical in traffic control strategies and traveler information systems. Predicting evolving traffic patterns on a road network is not a trivial task, and researchers have used advanced models to approximate traffic behavior. Over the years, these models include time series methods such as the autoregressive integrated moving average (ARIMA) model \cite{ahmed1979analysis, arima1988, ma2015nonlinear}, non-parametric regression models such as the k-nearest neighbor \cite{davis1991nonparametric, cai2016spatiotemporal} or support vector regression \cite{wu2004travel}, standard artificial neural network models of fully-connected \cite{dougherty1997short, abdulhai2002short} and recurrent neural networks \cite{tian2015predicting}. However, most recently, the state-of-the-art traffic prediction models are the graph convolutional neural network (GCNN) methods \cite{li2017diffusion, wang2018}. 

This paper first describes the graph convolution perspective and then develops a taxonomy of GCNN short-term traffic prediction models based on their components. Afterwards, we explore different variations along these components and eventually arrive at a variant that is similar to a traditional recurrent neural network. We then revisit a regression view of short-term traffic prediction using random forests, a powerful ensemble regression method. Finally, we compare the performance of these models using data from traffic simulations as well as the real world.

\subsection{Problem Definition}
Short-term traffic prediction can be performed at different levels, from the behaviors of individual vehicles to the traffic states of entire districts. This work investigates the prediction of traffic speed or flow at the level of individual links (segments of roads). Throughout this paper, we refer to road links as nodes, and connections between road links (such as road intersections) as edges in the context of GCNNs. In practice, a variety of factors such as weather and road design would also influence traffic patterns; however, similar to other works on this topic, we only use the past observations and the graph structure of the road network as inputs to our models.

The following notations are used throughout this paper:
\begin{itemize}
    \item $\mathcal{G} = (\mathcal{V}, \mathcal{E})$: The directed graph which describes the road network. $\mathcal{V}$ is the set of nodes, and $|\mathcal{V}| = N$. $\mathcal{E}$ is the set of connections or edges.
    \item $\mathcal{N}(i)$: The set of nodes in the neighborhood of node $i$. This is not restricted to the immediate neighbors of node $i$, and also includes node $i$ itself.
    \item $\mathbf{x}_i^{(t)}$ : A vector with length $d$ that represents the observation of node $i$ at time $t$.
    \item $\mathbf{X}^{(t)}$ : A matrix with size $(N \times d)$ that represents the observation of the entire road network at time $t$.
    \item $\hat{x}_i^{(t)}$ : A scalar that represents the prediction of node $i$ at time $t$. 
    \item $\hat{\mathbf{X}}^{(t)}$ : A vector with size $(N \times 1)$ that represents the prediction of the entire road network at time $t$.
    \item $H$: The prediction horizon.
\end{itemize}
Additionally, the variable $i$ and $j$ are reserved to distinguish a node and a time slice, respectively.

Using the above notation, we can define the prediction problem as learning a function $f$ that maps the past observations to predictions using the graph $\mathcal{G}$ and minimize the prediction error $L$ as follows:

\begin{equation}
    \hat{\mathbf{X}}^{(t+1)}, \hat{\mathbf{X}}^{(t+2)}, ..., \hat{\mathbf{X}}^{(t+H)} = f\left(\mathbf{X}^{(t)}, \mathbf{X}^{(t-1)}, ..., \mathcal{G}\right)
\label{prediction}
\end{equation}

\begin{equation}
    \min L = \sum_{i=1}^{|\mathcal{V}|} \sum_{j=1}^H \left\lVert\hat{x}_i^{(t+j)} - x_i^{(t+j)}\right\rVert
\label{loss}
\end{equation}

\subsection{The Graph Convolution Perspective}
A road network can be viewed as a graph, and road traffic is a dynamic process that develops gradually on the graph. Graph convolutional neural networks extend the notion of the convolution operation, which is commonly applied to analyzing visual imagery with a grid-like structure, to an operation that can be applied to graphs with arbitrary structures. Therefore, a GCNN is capable of extracting information using the spatial correlations between nodes in a graph and lends itself well to capturing the complex patterns needed for short-term traffic prediction. \cite{li2017diffusion} is the first application of graph convolutional neural network in short-term traffic prediction, and there have been many subsequent works that expand upon this idea.

In the GCNN perspective, every node has a defined neighborhood around it based on the structure of the graph. Using the formulation introduced by \cite{gilmer2017neural}, the operations of a GCNN can be divided into two phases, a message-passing phase and a readout phase. For a given node $i$, the message passing phase aggregates the information within its neighborhood $\mathcal{N}(i)$, then the subsequent readout phase applies the model parameters to create the output. In traffic, the predominant form of GCNN is the linear model introduced by \cite{kipf2016semi} and its extensions. In this framework, the message passing phase can be interpreted as taking a weighted sum of the inputs within the neighborhood, where the coefficients $a$ are determined by graph properties such as the adjacency matrix or the Laplacian matrix. Meanwhile, the readout phase is defined as a linear transformation with an activation function. Using the notation introduced earlier, this type of GCNN can be formulated as follows:
\begin{equation}
    \mathbf{k}_i = \rho\left(\sum\limits_{n \in \mathcal{N}(i)} a_{in}\mathbf{W}\mathbf{x}_n + \mathbf{b}\right)
\label{conv}
\end{equation}
where $\mathbf{k}_i$ is the output representation for node $i$, $\mathbf{x}_n$ is the graph input for node $n$, $a_{in}$ is the influence from node $n$ to node $i$ that is defined by the aggregation matrix, $\mathbf{W}$ and $\bf{b}$ are respectively the weight and bias terms of the model that transforms the input to hidden dimension, and $\rho(\cdot)$ denotes the activation function. 

In the more recent graph attention networks \cite{velivckovic2017graph}, $a_{in}$ are instead produced by an additional module that learns the relationship between every pair of nodes to assign weights for the aggregation. This can be achieved with a variety of attention mechanisms that exist in the literature such as the works of \cite{bahdanau2014neural, vaswani2017attention}, the mechanism cited in \cite{velivckovic2017graph} is as follows:
\begin{equation}
\begin{gathered}[t]
    a_{mn} = \text{softmax}_m\left(\text{LeakyReLU}\left(\boldsymbol{\alpha}^\top[\mathbf{W}\mathbf{x}_m \parallel \mathbf{W}\mathbf{x}_n]\right)\right) \\
    \mathbf{k}_i = \sigma\left(\sum\limits_{n \in \mathcal{N}(i)} a_{in}\mathbf{W}\mathbf{x}_n + \mathbf{b}\right) \\
\label{gat}
\end{gathered}
\end{equation}
where $\boldsymbol{\alpha}$ defines the linear layer that calculates the attention value between two nodes, and $\parallel$ denotes the concatenation operator. It is important to note that there is only one set of model weight $\mathbf{W}$ and bias $\mathbf{b}$ that is applied to all nodes in both formulations.

\subsection{Components of GCNN Traffic Prediction Models} \label{dims}
A short-term traffic prediction model can use GCNNs to capture the spatial correlations between different nodes of a road network; however, the model also needs to account for the changing dynamic of traffic through time. This is commonly achieved in the literature through the use of recurrent neural networks (RNNs) as exemplified by \cite{li2017diffusion, wang2018, bai2020adaptive}. An RNN is the predominant deep learning model for analyzing sequential data, consisting of repeated cells that form a temporal sequence. The output of each cell is used as the input to the next cell, and the parameters within the cell are shared across time steps. Consequently, this architecture can process sequential data of different lengths by varying the number of repetitions. The gated recurrent unit (GRU) \cite{cho2014learning} is a standard cell architecture used in RNNs, which operates as follows:
\begin{equation}
\begin{gathered}
    \mathbf{z}_i^{(t)} = \sigma\left(\mathbf{W}_z \mathbf{x}_i^{(t)} + \mathbf{U}_z\mathbf{k}_i^{(t-1)} + \mathbf{b}_z\right)\\
    \mathbf{r}_i^{(t)} = \sigma\left(\mathbf{W}_r\mathbf{x}_i^{(t)} + \mathbf{U}_r\mathbf{k}_i^{(t-1)} + \mathbf{b}_r\right)\\
    \mathbf{m}_i^{(t)} = \text{tanh}\left(\mathbf{W}_m\mathbf{x}_i^{(t)} + \mathbf{U}_m\left(\mathbf{r}_i^{(t)} * \mathbf{k}_i^{(t-1)}\right) + \mathbf{b}_m\right) \\
    \mathbf{k}_i^{(t)} = \left(1-\mathbf{z}_i^{(t)}\right) * \mathbf{k}_i^{(t-1)} + \mathbf{z}_i^{(t)} * \mathbf{m}_i^{(t)}
\label{gru}
\end{gathered}
\end{equation}
where $\mathbf{k}_i$ is the hidden states for node $i$; $\mathbf{z}_i, \mathbf{r}_i, \mathbf{m}_i$ are respectively the update gate, the reset gate, and the candidate hidden state; $*$ denotes the Hadamard product; $\sigma(\cdot)$ denotes the sigmoid function; while $\mathbf{W}$ and $\mathbf{U}$ represent the weights and $\mathbf{b}$ represents the biases of the model.

Typically, a short-term traffic prediction model based on GCNNs and RNNs integrates the two components by replacing the matrix multiplications in the GRU with a GCNN operation. For example, \eqref{gru} can be modified as follows:
\begin{equation}
\begin{aligned}
    \mathbf{z}_i^{(t)} &= \sigma\left(\sum\limits_{n \in \mathcal{N}(i)} a_{in}\left(\mathbf{W}_z\mathbf{x}_n^{(t)} + \mathbf{U}_z\mathbf{k}_n^{(t-1)}\right) + \mathbf{b}_z\right)\\
    \mathbf{r}_i^{(t)} &= \sigma\left(\sum\limits_{n \in \mathcal{N}(i)} a_{in}\left(\mathbf{W}_r\mathbf{x}_n^{(t)} + \mathbf{U}_r\mathbf{k}_n^{(t-1)}\right) + \mathbf{b}_r\right)\\
    \mathbf{m}_i^{(t)} &= \\ \text{tanh}&\left(\sum\limits_{n \in \mathcal{N}(i)} a_{in}\left(\mathbf{W}_m\mathbf{x}_n^{(t)} + \mathbf{U}_m\left(\mathbf{r}_i^{(t)} * \mathbf{k}_n^{(t-1)}\right)\right) + \mathbf{b}_m\right) \\
    \mathbf{k}_i^{(t)} &= \left(1-\mathbf{z}_i^{(t)}\right) * \mathbf{k}_i^{(t-1)} + \mathbf{z}_i^{(t)} * \mathbf{m}_i^{(t)}
\label{gc_gru}
\end{aligned}
\end{equation}

With this formulation, we developed a taxonomy of GCNN short-term traffic prediction models by identifying 3 GCNN components. The first 2 components are the operation concerning the input $\mathbf{x}_i^{(t)}$ and the last hidden state $\mathbf{k}_i^{(t-1)}$, which can be a standard matrix multiplication or a GCNN variant. The third component is the model weights $\mathbf{W}$, $\mathbf{U}$, and $\mathbf{b}$, which can be either shared among nodes or independent. We explore variations along these components in the next section. It is important to note that some works use other mechanisms to capture the temporal dynamics of traffic; however, this investigation is focused on RNN-based models due to their prevalence. 

\section{Methods}
\subsection{Variations on GCNN Components}
We begin with the work of \cite{wang2018}, which uses a standard matrix multiplication for input, convolution for the last hidden state, and shared model weights among nodes. This formulation transforms the first equation of \eqref{gru} to the following:
\begin{equation}
    \mathbf{z}_i^{(t)} = \sigma\left(\mathbf{W}_z \mathbf{x}_i^{(t)} + \sum\limits_{n \in \mathcal{N}(i)} a_{in} \mathbf{U}_z\mathbf{k}_n^{(t-1)} + \mathbf{b}_z\right)
\label{grnn}
\end{equation}
As in \eqref{gc_gru}, the remaining equations in \eqref{gru} can be transformed likewise and are omitted for brevity in this section.

We then experiment with changing the graph convolution operation to graph attention shown in \eqref{gat} and examine the different combinations of applying the attention operation to input and hidden states. For this investigation, we call this type of model the graph attention gated recurrent unit (GA-GRU) as it combines the concepts of graph attention networks and gated recurrent units. Equation \ref{ga_gru} is one configuration where attention is only applied to the last hidden state, i.e., GA-GRU (hidden).
\begin{equation}
\begin{gathered}
    a_{mn} = \frac{\text{exp}(\text{LeakyReLU}\left(\boldsymbol{\alpha}^\top[\mathbf{W}\mathbf{x}_m \parallel \mathbf{W}\mathbf{x}_n]\right))}
    {\sum\limits_{o \in \mathcal{N}(i)}\text{exp}(\text{LeakyReLU}\left(\boldsymbol{\alpha}^\top[\mathbf{W}\mathbf{x}_m \parallel \mathbf{W}\mathbf{x}_o]\right))} \\
    \mathbf{z}_i^{(t)} = \sigma\left(\mathbf{W}_z \mathbf{x}_i^{(t)} + \sum\limits_{n \in \mathcal{N}(i)} a_{in} \mathbf{U}_z\mathbf{k}_n^{(t-1)} + \mathbf{b}_z\right)\\
\label{ga_gru}
\end{gathered}
\end{equation}

    Afterwards, we explore the removal of shared weights among different nodes by designating a unique set of model weights $\mathbf{W}$, $\mathbf{U}$, and $\mathbf{b}$ for each node. In order to remove all weight sharing across nodes, we also replaced the shared attention layer in \eqref{ga_gru} with a trainable attention matrix. For this investigation, we call this type of model the attentional graph recurrent neural network, i.e., AGRNN (hidden). In this framework, \eqref{ga_gru} is transformed to the following:

\begin{equation}
    \mathbf{z}_i^{(t)} = \sigma\left(\mathbf{W}_{iz}\mathbf{x}_i^{(t)} + \sum\limits_{n \in \mathcal{N}(i)}a_{in}\mathbf{U}_{iz}\mathbf{k}_n^{(t-1)} + \mathbf{b}_{iz}\right) \\
\label{agrnn_h}
\end{equation}
In contrast with \eqref{ga_gru}, the subscript $i$ in all weights and biases signifies that each node contains its own model parameters, and the $a$ in this framework are learnable weights. 

\begin{figure}
    \centering
    \includegraphics[width=0.5\textwidth]{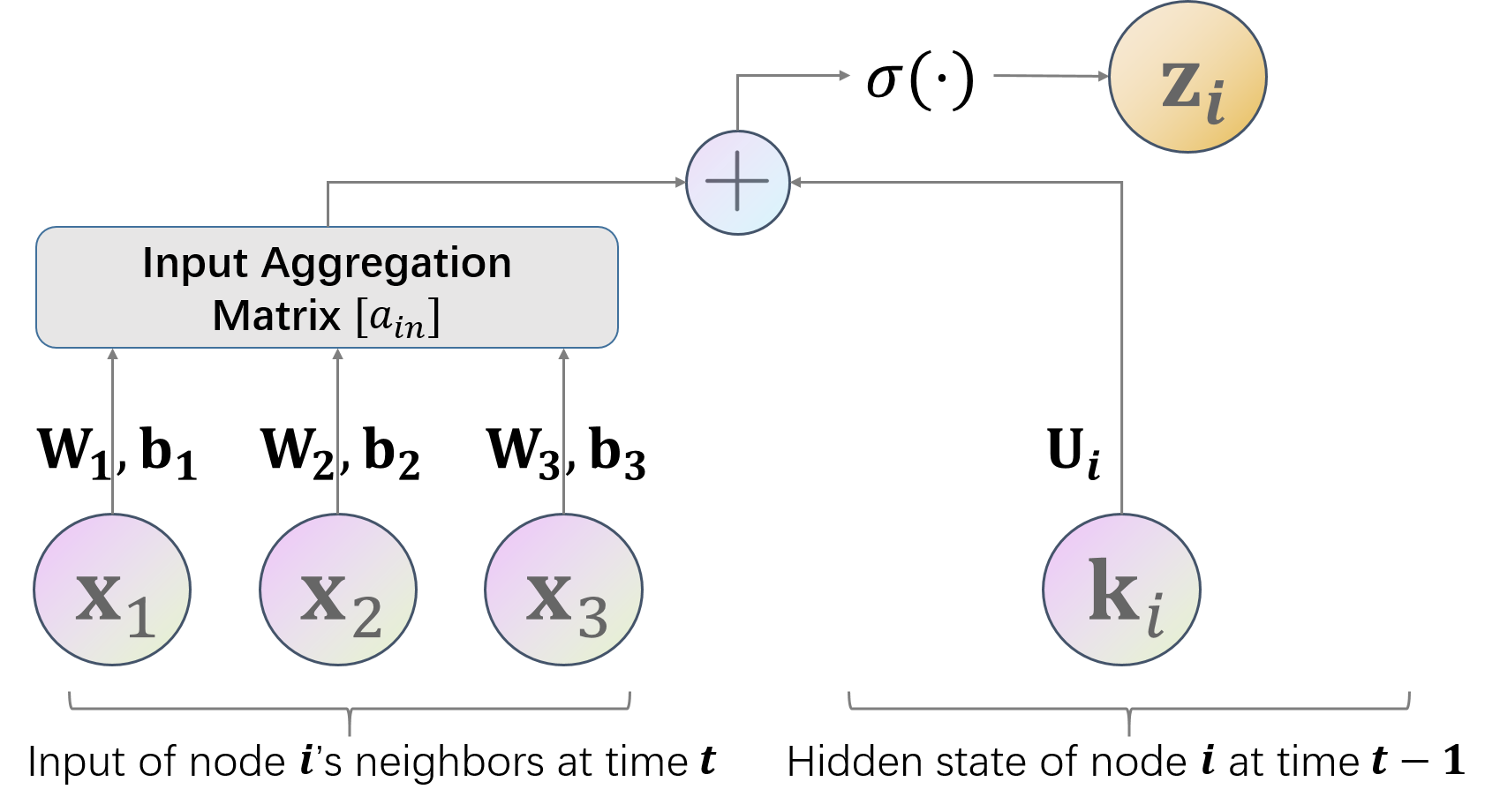}
    \caption{Input attention and independent input weights for the update gate of the GRU.}
    \label{fig:agrnn-input}
\end{figure}

We also highlight the input-only attention variant of the AGRNN, i.e., AGRNN (input). The diagram is shown in Fig. \ref{fig:agrnn-input}, and the equation is defined below:
\begin{equation}
\begin{gathered} 
    \mathbf{z}_i^{(t)} = \sigma\left(\sum\limits_{n \in \mathcal{N}(i)}a_{in}\mathbf{W}_{iz}\mathbf{x}_n^{(t)} + \mathbf{U}_{iz}\mathbf{k}_i^{(t-1)} + \mathbf{b}_{iz}\right) \\
\label{agrnn}
\end{gathered}
\end{equation}

Since there is no convolution or attention applied to the previous hidden states, the hidden states of different nodes do not influence one another. Combined with the independent model weights, this structure is akin to traditional recurrent neural networks with added location context and models of different nodes can be trained independently. 

Lastly, we also include in this comparison an example in the literature where the independent model weights are factorized according to the spatial correlations among nodes \cite{bai2020adaptive}, which results in a structure that shares weights between nodes yet applies a distinct model to every node. Table \ref{variations} summarizes the different variations of GCNN discussed in this section.

\begin{table}
\footnotesize
\rowcolors{2}{gray!25}{white}
\caption{The GCNN variants explored in this paper, categorized along the 3 component discussed in Section \ref{dims}}
\centering
\begin{tabular}{P{2.2cm}P{1.6cm}P{1.6cm}P{1.4cm}} 
\hline 
Model & Input & Hidden & Weights \\
\hline 
GRNN \cite{wang2018} & Multiplication & Convolution & Shared \\ 
GA-GRU (input) & Attention & Multiplication & Shared \\ 
GA-GRU (hidden) & Multiplication & Attention & Shared \\ 
GA-GRU (both) & Attention & Attention & Shared \\ 
AGRNN (input) & Attention & Multiplication & Independent \\ 
AGRNN (hidden) & Multiplication & Attention & Independent \\ 
AGRNN (both) & Attention & Attention & Independent \\
AGCRN \cite{bai2020adaptive} & Attention & Attention & Factorized \\ 
\hline
\end{tabular}
\label{variations}
\end{table}

\subsection{Random Forests}
Traffic propagation within a single time step is limited in space since the traffic state at a given link is independent of recent traffic states of faraway links. Although the exact radius of influence is changing and unknown, we can predict short-term traffic at a given link using recent observation within a neighborhood. We can then define short-term traffic prediction as a regression problem with the predictions ($\hat{x}_i^{(t+1)}, \hat{x}_i^{(t+2)}, ..., \hat{x}_i^{(t+H)}$) as the regressands, and the recent observations ($\mathbf{x}_n^{(t)}, \mathbf{x}_n^{(t-1)}, ... \forall n \in \mathcal{N}(i)$) as the regressors to facilitate the regression analysis.

In this work, we solve this regression problem by employing random forests \cite{breiman2001random}, which is a type of ensemble regression trees. A regression tree splits the input samples recursively into a tree-like decision diagram until it reaches the desired number of depth or leaf nodes. Each internal node of the tree contains a rule that splits the samples according to the value of a regressor and passes each split to the corresponding children node. Each leaf node of the tree contains a simple model that describes only the samples within its split. During prediction, we traverse the tree based on the regressors until we reach a leaf node, then the model within the node can generate the predicted regressand. With a large number of nodes, this approach can approximate complex functions with relatively simple models; however, this can also lead to overfitting. Random forests combats overfitting by splitting the training data to create multiple regression trees and averaging the output of all trees during prediction, which leads to a more robust regression model.


In the recent deep learning-focused traffic prediction works, we found a glaring lack of direct comparison against ensemble regression tree methods. This type of model consistently performs well in a variety of predictive problems, and we believe that it should not be overlooked in the short-term traffic prediction context. In contrast to neural network models, regression trees are much simpler to interpret and require minimal data preparation and model selection procedures. We used scikit-learn \cite{scikit-learn} to build and train the random forests model for this paper.

\section{Experimental Setup}
\subsection{Datasets}
\begin{figure}
    \centering
    \includegraphics[height=0.25\textwidth]{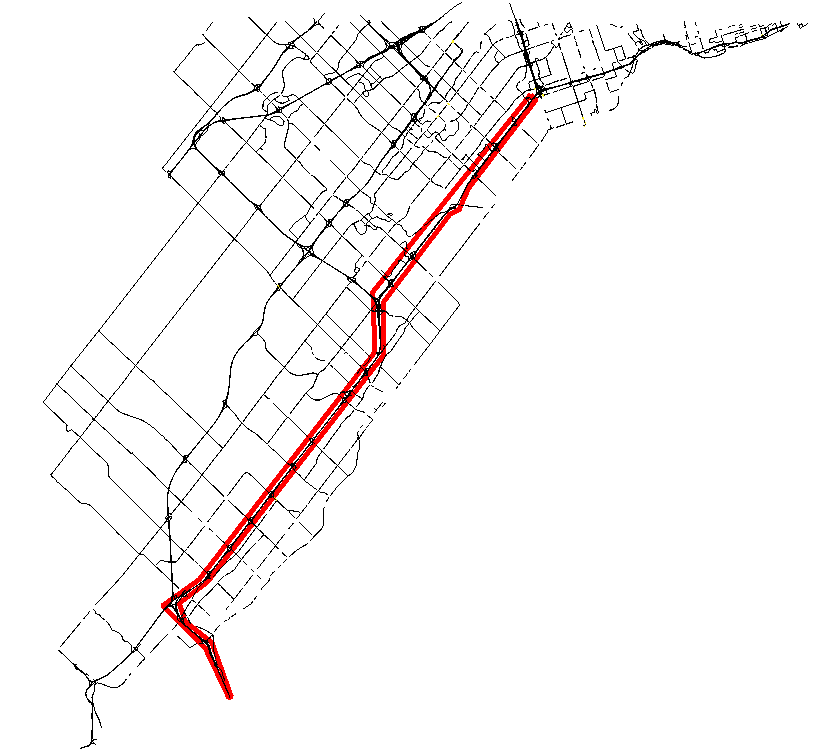}
    \caption{Map of the highway region chosen for this study.}
    \label{fig:qew}
\end{figure}

\begin{figure}
    \centering
    \includegraphics[height=0.25\textwidth]{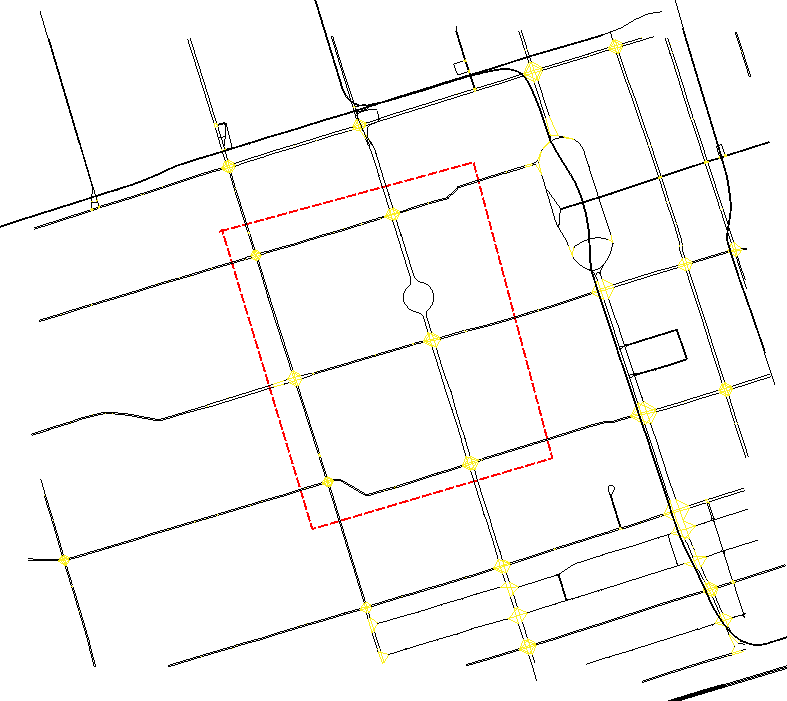}
    \caption{Map of the urban region chosen for this study.}
    \label{fig:downtown}
\end{figure}

\textbf{Toronto datasets}. We created two sets of data using  Aimsun Next \cite{aimsun} traffic simulation software. The first dataset is the traffic flow and speed of Queen Elizabeth Way, a highway in Ontario, Canada with 56 measurement links. The second dataset is the traffic flow and speed of downtown Toronto, Canada with 165 measurement links. The two regions are shown in Fig. \ref{fig:qew} and Fig. \ref{fig:downtown}.


For the simulation, we used the travel demand collected from a survey in 2016 \cite{ashby2018transportation}, and further calibrated using measurements from the loop detectors installed along the roads \cite{kamel2019integrated}. We built the simulation model using morning peak-hour travel demands, and each simulation is for the 4-hour period between 6:00 and 10:00 AM. The speed (distance traveled per unit time) and flow (number of vehicles per unit time) for every link were extracted from the simulations in 1-minute interval. Speed and flow were selected because they are the most common form of data in the real-world measured using loop detectors and the GPS. To augment the data, the simulation was run 50 times with the original travel demands multiplied by a random scalar factor between 0.5 and 1.5.

\textbf{California datasets}. The PeMS04 and PeMS08 datasets are collected from the Caltrans Performance Measurement System (PeMS) of districts in California. The PeMS04 dataset consists of 307 measurement locations on highways surrounding San Jose, California, dating from January 1st to February 28th in 2018. The PeMS08 dataset consists of 170 measurement locations on highways surrounding San Bernardino, California, dating from July 1st to August 31st in 2016. For both datasets, the measurement interval is 5 minutes, which corresponds to 288 data points per day. The adjacency matrix is defined according to road distance and connectivity. We follow the evaluation procedure established by other papers \cite{bai2020adaptive, mengzhang2020spatial}, which uses the last 12 observations to predict the next 12 time steps, i.e., use the past hour of traffic data to predict that of the next hour.

\subsection{Model Selection}
We evaluated our model against a selection of different methods that are listed below, including other graph convolutional neural networks as well as time series analysis methods. We built all neural network models, including GA-GRUs and AGRNNs, using PyTorch \cite{paszke2019pytorch} in this paper. Additionally, we tuned the hyperparameters summarized in Table \ref{tab:hyperparameter} using coordinate descent.

\begin{table*}[!ht]
\centering
\rowcolors{2}{gray!25}{white}
\caption{Hyperparameter selection}
    \begin{tabular}{c c c}
        \hline
        Name & Range & Applicable Models \\
        \hline
        Neighborhood size (number of steps to target node) & \{1, 2, 3, 4, 5, 6\} & All models except ARIMA\tabularnewline
        Number of historical time steps & \{2, 3, ..., 9\} & All models on simulation datasets \tabularnewline
        Number of historical time steps & 12 & All models on PeMS datasets \tabularnewline
        Number of hidden features & \{32, 64, 128\} & Neural network models \tabularnewline
        Batch size & \{32, 64, 128\} & Neural network models \tabularnewline
        Learning rate & \{1e-5, 1e-4, 1e-3\} & Neural network models \tabularnewline
        L2 regularization strength & \{1e-5, 1e-4, 1e-3\} & Neural network models \tabularnewline
        Number of training epochs & \{50, 100, ..., 2000\} & Neural network models \tabularnewline
        Number of differences & \{0, 1, 2\} & ARIMA \tabularnewline
        Number of trees & \{100, 150, 200\} & Random forests \tabularnewline
        Tree depth & \{5, 10, 15, 20\} & Random forests \tabularnewline
        \hline
    \end{tabular}
\label{tab:hyperparameter}
\end{table*}

\begin{table*}
\centering
\footnotesize
\rowcolors{3}{white}{gray!25}
\caption{Performance comparison of traffic speed prediction models for simulated highway dataset}
\begin{tabular}{>{\centering}p{2.2cm} | >{\centering}p{1cm}  >{\centering}p{1cm} >{\centering}p{1cm} | >{\centering}p{1cm}  >{\centering}p{1cm} >{\centering}p{1cm}} 
\hline
& \multicolumn{3} {c}{5-minute horizon} & \multicolumn{3} {c}{15-minute horizon} \tabularnewline
\hline
Model & MAE (km/h) & MAPE (\%) & RMSE (km/h) & MAE (km/h) & MAPE (\%) & RMSE (km/h) \tabularnewline
\hline
Historical Average & \multicolumn{3} {|c|}{Not applicable} & \multicolumn{3} {c}{Not applicable} \tabularnewline
ARIMA & 4.41 & 11.96 & 9.13 & 7.52 & 20.73 & 15.47 \tabularnewline
GCN & 3.72 & 9.05 & 5.95 & 5.24 & 13.08 & 8.99 \tabularnewline
GRNN & \underline{3.24} & 9.66 & \underline{5.31} & 5.18 & 14.90 & 9.00 \tabularnewline
GA-GRU (input) & 4.99 & 15.30 & 7.72 & 13.09 & 38.09 & 16.83 \tabularnewline
GA-GRU (hidden) & 3.38 & 10.30 & 5.35 & 5.00 & 15.31 & 8.55 \tabularnewline
GA-GRU (both) & 4.08 & 12.55 & 6.32 & 13.48 & 47.70 & 16.95 \tabularnewline
AGRNN (input) & 3.28 & 9.71 & 5.46 & 4.90 & 14.85 & 8.63 \tabularnewline
AGRNN (hidden) & 3.92 & 9.14 & 6.00 & 4.99 & 12.13 & 8.12 \tabularnewline
AGRNN (both) & 3.53 & 8.69 & 6.17 & \underline{3.84} & \textbf{9.52} & \underline{6.86} \tabularnewline
AGCRN & 3.41 & \textbf{7.70} & 7.38 & 4.28 & \underline{10.00} & 8.84 \tabularnewline
Random forests & \textbf{2.77} & \underline{8.02} & \textbf{4.73} & \textbf{3.60} & 10.31 & \textbf{6.51} \tabularnewline
\hline
\end{tabular}
\label{res:highway}
\end{table*}

\begin{itemize}
\renewcommand\labelitemi{--}
    \item \textbf{Historical Average}: A time series model that predicts the average of observations from the same time of day in previous weeks. It is not applicable to simulated datasets since the simulation model simulates only one day and has no long-term time series.
    \item \textbf{ARIMA}: A time series model that is well-documented in the traffic prediction literature. This is a univariate model so only the recent observations at the same link are used to generate the prediction. We used pmdarima \cite{pmdarima} to construct the ARIMA model for this paper. 
    \item \textbf{GCN} \cite{kipf2016semi}: A Graph Convolutional Network that contains 1 hidden layer, and the GCN output layer is connected with a fully-connected layer to predict traffic states.
    \item \textbf{AGCRN}: This model \cite{bai2020adaptive} leverages node embeddings to learn adaptive spatial correlations, and uses node-specific parameters for convolution. To capture the temporal correlations, gated recurrent units are adopted.

\end{itemize}

\subsection{Evaluation Metrics}
To assess the performance, we selected three of the most commonly used time series regression metrics: mean absolute error (MAE), mean absolute percentage error (MAPE), and root-mean-square error (RMSE). MAE is the average of the absolute error across all predictions while MAPE is the average of the absolute relative error that emphasizes lower values. Meanwhile, RMSE is the square root of the average squared errors, which is also the standard deviation of all prediction errors. For a prediction horizon $H$, given predictions ${\hat{x}_i^{(t+1)}, \hat{x}_i^{(t+2)}, ...,  \hat{x}_i^{(t+H)}}$ and the actual observed value ${x_i^{(t+1)}, x_i^{(t+2)}, ..., x_i^{(t+H)}}$, we can calculate the three metrics using the equations below:

\begin{equation}
    \text{MAE} =\frac{1}{|\mathcal{V}|} \frac{1}{H} \sum\limits_{i=1}^{|\mathcal{V}|} \sum\limits_{j=1}^{H} \left|x_i^{(t+j)} - \hat{x}_i^{(t+j)}\right|
\label{mae}
\end{equation}

\begin{equation}
    \text{MAPE} =\frac{1}{|\mathcal{V}|} \frac{1}{H} \sum\limits_{i=1}^{|\mathcal{V}|} \sum\limits_{j=1}^{H} \left|\frac{x_i^{(t+j)} - \hat{x}_i^{(t+j)}}{x_i^{(t+j)}}\right|
\label{mape}
\end{equation}

\begin{equation}
    \text{RMSE} = \sqrt{\frac{1}{|\mathcal{V}|} \frac{1}{H} \sum\limits_{i=1}^{|\mathcal{V}|} \sum\limits_{j=1}^{H} \left(x_i^{(t+j)} - \hat{x}_i^{(t+j)}\right)^2}
\label{rmse}
\end{equation}

For the Toronto datasets, we used both 5th minute and 15th minute as the prediction horizon and computed error using only the prediction at the horizon $\hat{x}_i^{(t+H)}$. Meanwhile, on the California datasets, we followed the convention of \cite{bai2020adaptive, mengzhang2020spatial} and computed error using all predictions up to the prediction horizon $H$, i.e., ${\hat{x}_i^{(t+1)}, \hat{x}_i^{(t+2)}, ...,  \hat{x}_i^{(t+H)}}$. 

Although the above metrics conveniently produce numerical values for easy comparison across models, they are unable to represent all model aspects.  Therefore, we also measured the complexity of each model for a more well-rounded comparison as shown in Table \ref{res:complexity}. 

\section{Results and Discussion}
We performed the evaluation using a data split of 60\% training, 20\% validation, and 20\% testing for each dataset. We then recorded each metric to produce results shown in Tables \ref{res:highway}, \ref{res:urban}, and \ref{res:pems} below, where the bolded number is the lowest error and the underlined number is the second lowest error. In addition, we also report the model complexity for the 5-minute prediction horizon on the highway dataset in Table \ref{res:complexity}.

\begin{table*}
\centering
\footnotesize
\rowcolors{3}{white}{gray!25}
\caption{Performance comparison of traffic speed prediction models for simulated urban dataset}
\begin{tabular}{>{\centering}p{2.2cm} | >{\centering}p{1cm}  >{\centering}p{1cm} >{\centering}p{1cm} | >{\centering}p{1cm}  >{\centering}p{1cm} >{\centering}p{1cm}} 
\hline
& \multicolumn{3} {c}{5-minute horizon} & \multicolumn{3} {c}{15-minute horizon} \tabularnewline
\hline
Model & MAE (km/h) & MAPE (\%) & RMSE (km/h) & MAE (km/h) & MAPE (\%) & RMSE (km/h) \tabularnewline
\hline
Historical Average & \multicolumn{3} {c|}{Not applicable} & \multicolumn{3} {c}{Not applicable} \tabularnewline
ARIMA & 4.87 & 30.98 & 7.73 & 5.43 & 35.12 & 8.58 \tabularnewline
GCN & 4.30 & 24.33 & \underline{6.53} & 4.62 & 25.21 & 6.94 \tabularnewline
GRNN & 4.37 & 24.88 & 7.06 & 4.94 & 29.32 & 7.78 \tabularnewline
GA-GRU (input) & 4.85 & 31.06 & 7.65 & 5.31 & 33.45 & 8.39 \tabularnewline
GA-GRU (hidden) & 4.24 & 24.04 & 6.91 & 4.83 & 28.48 & 7.98\tabularnewline
GA-GRU (both) & 4.62 & 29.27 & 7.40 & 4.96 & 31.18 & 7.90 \tabularnewline
AGRNN (input) & 4.11 & 22.49 & 6.88 & 4.31 & 24.44 & 7.21 \tabularnewline
AGRNN (hidden) & 4.23 & 24.55 & 6.71 & 4.43 & 25.77 & 7.02 \tabularnewline
AGRNN (both) & 4.08 & \underline{22.35} & 6.62 & \underline{4.27} & \textbf{23.88} & \underline{7.02} \tabularnewline
AGCRN & \underline{4.02} & 23.59 & 7.02 & 4.49 & 28.40 & 7.78 \tabularnewline
Random forests & \textbf{3.89} & \textbf{22.18} & \textbf{6.21} & \textbf{4.17} & \underline{24.37} & \textbf{6.66} \tabularnewline
\hline
\end{tabular}
\label{res:urban}
\end{table*}

\begin{table*}[!htbp]
\centering
\footnotesize
\rowcolors{3}{white}{gray!25}
\caption{Performance comparison of traffic flow prediction models on PeMS04 and PeMS08 dataset}
\begin{tabular}{>{\centering}p{2.2cm} | >{\centering}p{1cm}  >{\centering}p{1cm} >{\centering}p{1cm} | >{\centering}p{1cm}  >{\centering}p{1cm} >{\centering}p{1cm}} 
\hline
& \multicolumn{3} {c}{PeMS04} & \multicolumn{3} {c}{PeMS08} \tabularnewline
\hline
Model & MAE & MAPE (\%) & RMSE & MAE & MAPE (\%) & RMSE  \tabularnewline
\hline
Historical Average & 24.99 & 16.07 & 41.84 & 21.21 & 13.72 & 36.73 \tabularnewline
ARIMA & 27.53 & 20.55 & 42.44 & 22.67 & 14.92 & 35.08 \tabularnewline
GCN & 23.72 & 17.92 & 37.47 & 21.09 & 14.42 & 31.45 \tabularnewline
GRNN & 30.66 & 25.02 & 46.06 & 26.09 & 21.92 & 39.07 \tabularnewline
GA-GRU (input) &32.78 &30.24 &46.65 & 27.73 & 42.20 & 38.78 \tabularnewline
GA-GRU (hidden) &24.73 & 17.17 & 38.18 & 19.89 & 14.01 & 30.73 \tabularnewline
GA-GRU (both) &29.35 &25.23 &42.74 &23.53 &19.96 &34.46 \tabularnewline
AGRNN (input) & 24.01 & 17.37 & 37.82 & 21.04 & 14.57 & 31.19 \tabularnewline
AGRNN (hidden) & 22.97 & 16.32 & 37.25 & 20.31 & 13.48 & 30.80 \tabularnewline
AGRNN (both) & 23.77 & 16.54 & 38.53 & 22.04 & 14.33 & 34.16 \tabularnewline
AGCRN & \textbf{19.86} & \textbf{13.06} & \textbf{32.57} & \textbf{16.08} & \textbf{10.40} & \textbf{25.55} \tabularnewline
Random forests & \underline{20.74} & \underline{13.76} & \underline{35.03} & \underline{16.64} & \underline{10.95} & \underline{26.95} \tabularnewline
\hline
\end{tabular}
\label{res:pems}
\end{table*}

\begin{table}[!ht]
\centering
\footnotesize
\rowcolors{2}{white}{gray!25}
\captionof{table}{Model complexity measured in number of parameters}
\begin{threeparttable}
\renewcommand{\TPTminimum}{\linewidth}
\makebox[\linewidth]{%
\begin{tabular}{>{\centering}p{3cm}  >{\centering}p{1.5cm} } 
\hline
Model & Complexity \tabularnewline
\hline
Historical Average & 48960 \tabularnewline
ARIMA & 1035 \tabularnewline
GCN & 66587944 \tabularnewline
AGCRN & 150112 \tabularnewline
GRNN & 3393 \tabularnewline
GA-GRU (input) & 1025 \tabularnewline
GA-GRU (hidden) & 1025 \tabularnewline
GA-GRU (both) & 1121 \tabularnewline
AGRNN (input) & 8602510 \tabularnewline
AGRNN (hidden) & 605710 \tabularnewline
AGRNN (both) & 634610 \tabularnewline
Random forests\tnote{a} & 171195284 \tabularnewline
\hline
\end{tabular}}
\begin{tablenotes}[flushleft]
\item[a] It is difficult to quantify the exact number of parameters in a random forests model (since tree structures are semi-parametric), therefore we instead report the combined number of nodes across all trees. The actual number of parameters is at least several times this value.
\end{tablenotes}
\end{threeparttable}
\label{res:complexity}
\end{table}

Across all experiments, performances improve from that of GRNN, GA-GRUs, to AGRNNs. First, this indicates that using an attention mechanism to learn spatial correlations is better than using a fixed adjacency matrix. Besides, the node-specific convolutional weights can capture distinct traffic pattern in each node and improve accuracy. Moreover, the AGRNN (input) model is competitive with other GCNNs according to all error metrics; which signifies that the propagation of hidden states among nodes between consecutive RNN time steps is not essential in achieving accurate prediction. It should be noted that although our experiments keep the other components of the model constant among GRNN, GA-GRUs, and AGRNNs, the findings of this work may not generalize to other architectures, such as multiple graph convolutional layers or different RNN configurations.

The AGRNN with input-only attention is an extreme version of separating GCNN weights to create independent models; meanwhile, the factorized weights in AGCRN can be viewed as a tradeoff between having completely shared weights of GCNNs and completely independent weights. The superior results of the AGCRN model in our experiments suggest that this tradeoff approach is worthy of further investigation.

The experiment results also show that the random forests model exhibits the lowest error on the simulated datasets, while being narrowly outperformed by the AGCRN model on the California datasets. This supports the hypothesis that short-term traffic prediction can be formed as a regression problem and further highlights that sharing model weights and latent states are inconsequential in attaining model accuracy. However, the random forests model contains by far the largest number of parameters across all experiments.
Overall, the result suggests that while GCNNs can be more compact, random forests regression remains competitive and should not be overlooked in short-term traffic prediction.

\section{Conclusion and Future Work}
This work classifies and evaluates the variants of GCNNs for short-term traffic prediction based on their components. We found that incorporating attention, matrix factorization, and location-specific model weights are beneficial to overall performance. However, the traditional random forest regression method cannot be ignored for its excellent performance compared to that of the latest GCNN models. In terms of future work, we can expand this comparison to other components and encompass a larger selection of recent works. Furthermore, we plan to harness these findings to improve existing deep learning traffic prediction methods.


\bibliography{refs}

\end{document}